\documentclass[journal]{IEEEtran}
\usepackage{rotating}
\usepackage{times}
\usepackage{epsfig}
\usepackage{graphicx}
\usepackage{amsmath}
\usepackage{amssymb}
\usepackage{picins}
\usepackage{epstopdf}
\usepackage[center]{subfigure}
\usepackage{mathrsfs}

\usepackage{cite}
\usepackage{algorithm}
\usepackage{algorithmic}
\usepackage{multirow}
\usepackage{color,hyperref}
\usepackage{url}

\newtheorem{theorem}{Theorem}
\newtheorem{lemma}{Lemma}

\newtheorem{proof}{Proof}

\newcommand{\argmin}{\mathop{{\rm arg}\min}}

\newcommand{\RNum}[1]{\uppercase\expandafter{\romannumeral #1\relax}}

\definecolor{darkblue}{rgb}{0.0,0.0,1.0}
\hypersetup{colorlinks,breaklinks,
	linkcolor=darkblue,urlcolor=darkblue,
	anchorcolor=darkblue,citecolor=darkblue}

\makeatletter
\long\def\@makecaption#1#2{\ifx\@captype\@IEEEtablestring%
	\footnotesize\begin{center}{\normalfont\footnotesize #1}\\
		{\normalfont\footnotesize\scshape #2}\end{center}%
	\@IEEEtablecaptionsepspace
	\else
	\@IEEEfigurecaptionsepspace
	\setbox\@tempboxa\hbox{\normalfont\footnotesize {#1.}~~ #2}%
	\ifdim \wd\@tempboxa >\hsize%
	\setbox\@tempboxa\hbox{\normalfont\footnotesize {#1.}~~ }%
	\parbox[t]{\hsize}{\normalfont\footnotesize \noindent\unhbox\@tempboxa#2}%
	\else
	\hbox to\hsize{\normalfont\footnotesize\hfil\box\@tempboxa\hfil}\fi\fi}
\makeatother

\begin{document}
	%
	\title{Sparse Generalized Canonical Correlation Analysis: Distributed Alternating Iteration based Approach}
	
	%
	%
	\author{Jia~Cai, Kexin~Lv, Junyi~Huo, Xiaolin~Huang~\IEEEmembership{Senior Member,~IEEE}, Jie~Yang
		\thanks{The work described in this paper is supported partially by  National Natural Science Foundation of China (11871167, 11671171, 61977046),  Science and Technology Program of Guangzhou (201707010228),  Special Support Plan for High-Level Talents of Guangdong Province (2019TQ05X571), Project of Collaborative Innovation Development Center of Pearl River Delta Science \& Technology Finance Industry (19XT01), Foundation of Guangdong Educational Committee (2019KZDZX1023). The corresponding author is Xiaolin Huang.}
		\thanks{J.~Cai is with  School of Statistics and Mathematics, Guangdong University  of Finance $\&$ Economics, also with Big Data and Educational Statistics Application Laboratory, 21 Chisha Road, Guangzhou  510320, Guangdong, P. R. China. Email:~jiacai1999@gdufe.edu.cn.}
		
		\thanks{K.~Lv, X.~Huang and J.~Yang are with Institute of Image Processing and Pattern Recognition, Shanghai Jiao Tong University, also with the MOE Key Laboratory of System
			Control and Information Processing, 800 Dongchuan Road, Shanghai, 200240, P.R. China. Emails: \{kelen\_lv, xiaolinhuang, jieyang\}@sjtu.edu.cn.}
		
		\thanks{Junyi Huo is with School of Electronics and Computer Science, University of Southampton, University Road, Southampton,  SO17 1BJ, United Kingdom. Email:~jh4a19@soton.ac.uk.}
		
	}
	
	%
	%

	\markboth{}%
	{****}
	%



	\maketitle
	\begin{abstract}
Sparse canonical correlation analysis (CCA) is a useful statistical tool to detect latent information with sparse structures. However, sparse CCA works only for two datasets, i.e., there are only two views or two distinct objects. To overcome this limitation, in this paper, we propose a sparse generalized canonical correlation analysis (GCCA), which could detect the latent relations of multiview data with sparse structures. Moreover, the introduced sparsity could be considered as Laplace prior on the canonical variates. Specifically, we convert the GCCA into a linear system of  equations and impose $\ell_1$ minimization penalty for sparsity pursuit. This results in a nonconvex problem on Stiefel manifold, which is difficult to solve.  Motivated by Boyd's consensus problem,  an algorithm based on   distributed alternating iteration approach is developed and theoretical consistency analysis is investigated elaborately under mild conditions.  Experiments on several synthetic and real world datasets demonstrate the effectiveness of the proposed algorithm.
	\end{abstract}


	%
	\IEEEpeerreviewmaketitle

	\section{Introduction}\label{sec:intr}
	\IEEEPARstart{C}{anonical} correlation analysis (CCA), launched by \cite{hotelling1936relations}, is a celebrated statistical tool for finding the correlation  between two sets of multidimensional variables. The two sets of variables can be considered as two views of one object or a view of two distinct objects. The main aim of CCA is to find  two sets of canonical variables (weight vectors) such that the projected variables in the lower-dimensional space are maximally correlated. Due to its efficiency of finding latent information,  CCA has been widely used in many branches of signal processing and data analytics, including but not limited to, cross language document retrieval \cite{vinokourov2003inferring}, genomic data analysis \cite{yamanishi2003extraction}, and functional magnetic resonance imaging \cite{friman2001detection}. Both theoretical  and algorithmic analysis of CCA have been widely investigated, see, e.g., \cite{michaeli2016nonparametric, gao2017stochastic, cai2017constrained, zhang2017robust, tan2018sparse, ma2015finding, wang2015stochastic, ge2016efficient}.
	
However, a large portion of features are not informative for high-dimensional data in the field of data analysis. When the canonical variables involve all features in the original space, the canonical variates are usually not sparse. to interpret canonical variables in high-dimensional data analysis, sparsity is often introduced. Similarly, since the establishment of compressive sensing, sparsity has been found to be efficient in enhancing the performance of many learning methods, if a suitable sparse structure could be found. In the field of CCA, there have been many efforts to impose sparsity, which could  not be obtained by the original CCA. For instance, the sparse penalized CCA algorithm \cite{waaijenborg2008quantifying}, the penalized matrix decomposition approach based sparse CCA method  \cite{witten2009penalized}, and the sparse CCA under primal-dual framework  \cite{hardoon2011sparse} have been developed.  \cite{chu2013sparse} selected the sparsest CCA solution from a subset of all solutions via the linearized Bregman method.  \cite{chen2013sparse} developed  a precision adjusted iteration thresholding method to estimate the sparse canonical weights, while  \cite{gao2017sparse}  investigated a two stage based sparse CCA method, where the first initialization stage was solved by Alternating Direction Method of Multipliers (ADMM), and then a group-Lasso based method was utilized to find the sparse weights in the second refinement stage.
	
In spite of great success, CCA and sparse CCA can only handle two datasets, which heavily limits the applications on multiview analysis and multi-modal learning. To overcome this problem, generalized CCA (GCCA) methods have been proposed. Among several attempts, tensor CCA \cite{luo2015tensor}, GCCA \cite{kang2013sparse}, weighted  GCCA \cite{benton2016learning},   scalable MAX-VAR GCCA \cite{fu2017scalable} and Deep GCCA (DGCCA, \cite{benton})  have shown good performance to deal with multiple datasets. Similarly to CCA, suitably imposing sparsity on GCCA could improve the performance. But the sparsity pursuit method designed for CCA can not be readily extended to GCCA.

Simply coping the technique from CCA to sparse CCA is not applicable for GCCA. To the best of our knowledge, only \cite{kang2013sparse} and \cite{kanatsoulis2018structured} discussed   sparse GCCA  methods.  Kang et al. \cite{kang2013sparse} designed a sparse GCCA under the special constraints that the data matrices and the projected variables have multiple regression relationships, while Kanatsoulis et al. \cite{kanatsoulis2018structured} discussed a primal-dual decomposition based ADMM GCCA approach for large-scale problems. Theoretical convergence is guaranteed by introducing Robinson's condition,  which requires  the number of canonical components should far less than the number of samples or features.
However, the performance of sparse GCCA is far from satisfactory in both sparsity and accuracy. The aim of this paper is to establish a sparse GCCA method.  The contributions of this paper are summarized as follows.
\begin{itemize}
		\item We formulate GCCA into the form of linear  system of equations, which serves as the basis for imposing sparsity. This leads to a nonconvex problem on Stiefel manifold.
		\item Based on the developed GCCA related equations and the model demonstrated in \cite{via2007learning},  we elegantly develop a novel sparse GCCA algorithm using augmented distributed alternative iteration method, which is a generalization of the consensus problem in \cite{Boyd2011}.
		\item Theoretical consistency of the proposed algorithm is judiciously investigated via convex analysis related theory under mild conditions.
 \item Experiments on gene data and Europarl dataset demonstrate the effective and efficiency of the proposed method.
\end{itemize}
The remainder of the paper is organized as follows. In Section \ref{section2}, we give a brief review of CCA and GCCA.  Section \ref{section3} devotes to the design of the new sparse GCCA and its solving algorithm. Section \ref{section4}  discusses the experimental results. Section \ref{conclu} concludes the paper. The proof of the main results go to the appendix.

	\section{Brief review of CCA and GCCA}\label{section2}
	\subsection{Canonical correlation analysis}
In this section, we briefly review canonical correlation analysis (CCA). Let $x\in R^{n_1}$ and $y\in R^{n_2}$ be two random variables. Denote $X=(x_1,\cdots,x_m)\in R^{n_1\times m}$, $Y=(y_1,\cdots,y_m)\in R^{n_2\times m}$. Without loss of generality, we assume both $\{x_i\}^m_{i=1}$ and $\{y_i\}^m_{i=1}$ have zero mean, i.e., $\sum^m_{i=1}x_i=0$ and $\sum^m_{i=1}y_i=0$. CCA solves the following problem
\begin{align}\label{cca}
	\max_{w1\neq 0, w_2\neq 0} & ~~  w^T_1 XY^T w_2\nonumber\\
	\mathrm{s.t.} & ~~   w^T_1XX^T w_1=1\nonumber\\
	& ~~ w^T_2YY^T w_2=1.
\end{align}
In Eq. (\ref{cca}), only one pair of canonical variables could be found. For more pairs,  \cite{chu2013sparse,hardoon2004canonical} extended CCA to the following multiple CCA,
\begin{align}\label{multiplecca}
	\max_{W_1, W_2} & ~~ {\rm Trace}(W^T_1 XY^T W_2)\nonumber\\
	\mathrm{s.t.} & ~~ W^T_1XX^T W_1= I_\ell,~~W_1\in \mathbb{R}^{n_1\times\ell}\nonumber\\
	& ~~  W^T_2YY^T W_2=I_\ell,~~W_2\in \mathbb{R}^{n_2\times\ell},
\end{align}
where $I_\ell$ denotes the $\ell\times \ell$ identity matrix, $\ell$ also stands for the number of columns of $W_i$ ($i=1,2$). When $\ell=1$, Eq. (\ref{multiplecca}) reduces to Eq. (\ref{cca}).  Obviously, both CCA and multiple CCA could only deal with two datasets.

\subsection{Generalized canonical correlation analysis}
To detect the relations of multiple multivariate datasets (more than two),  a generalized CCA that considers the sum of correlations was proposed by \cite{carroll1968equations}:
\begin{equation}\label{scorr_gcca}
\min_{W_i\neq 0, W_j\neq 0}  \sum^{J-1}_{i=1}\sum^J_{j=i+1}\|W^T_i X_i-W^T_jX_j\|^2_F,
\end{equation}
where $F$ denotes the Frobenius norm of a matrix, and $J$ stands for the number of views.  However, Eq. (\ref{scorr_gcca}) is a  NP-hard problem. To efficiently study the latent information of multiple datasets,  MAX-VAR formulation of GCCA was proposed \cite{kettenring1971canonical}:
\begin{align}\label{maxvar}
	&\min_{\{W_j\}^J_{j=1},G}   \sum^J_{j=1}\| G-W^T_j X_j\|^2_F\nonumber\\
	&s.t.\qquad   GG^T=I,
\end{align}
where $G\in \mathbb{R}^{\ell\times m} $ is a common latent representation of the different views.  In the literature, Eq. (\ref{maxvar}) is solved by selecting the principal eigenvectors of a matrix aggregated from the correlation matrix of different views, i.e., the rows of the optimal $G$ are the eigenvectors of the following matrix,
$$M=\sum_{j=1}^J X^T_j (X_jX^T_j)^{-1}X_j.$$
When $J=2$,  Eq. (\ref{maxvar}) becomes
\begin{align}\label{twoviewcca}
	&\min_{W_1, W_2,G}  \| G-W^T_1 X_1\|^2_F+\| G-W^T_2 X_2\|^2_F\nonumber\\
	&s.t.\qquad GG^T=I.
\end{align}
Recalling the  constraints in Eq. (\ref{multiplecca}), we have
\begin{align*}
	&~~~~ \|W^T_1X_1-W^T_2X_2\|^2_F\\
	&= {\rm Trace} (W^T_1X_1X^T_1W_1+W^T_2X_2X^T_2W_2\\
	&~~~~ -W^T_2X_2X_1^TW_1-W^T_1X_1X^T_2W_2)\\
	&=2\ell-2{\rm Trace} (W^T_1X_1X_2^TW_2).
\end{align*}
Combining it with the triangle inequality
$$\|W^T_1X_1-W^T_2X_2\|^2_F\leq 2\|W^T_1X_1-G\|^2_F+2\|G-W^T_2X_2\|^2_F,$$
one can find that the target function in Eq. (\ref{twoviewcca}) is a relaxation of that in Eq.  (\ref{multiplecca}).
	
\section{Sparse GCCA: Model and Algorithm}\label{section3}
\subsection{New formulation of sparse GCCA}
In this section, we will propose a sparse GCCA model and develop its solving algorithm. The basic idea is to convert problem (\ref{maxvar}) into a linear system of equations by considering the optimality conditions and employing singular value decomposition  (SVD) technique. Specifically, according to
\begin{align*}
	&\|G-W^T_jX_j\|^2_F\\
	&= {\rm Trace} (GG^T-W^T_j X_j G^T-GX^T_jW_j+W^T_j X_j X^T_j W_j).
\end{align*}
We can take the derivatives with respect to $W_j$ and obtain the optimality conditions as below,
$$X_jX^T_j W_j=X_jG^T.$$
Typically,  $X_j$ is  a low rank matrix and there is redundancy in the above equation. Denote the reduced SVD of $X_j$  as the following,
\begin{equation}\label{decomp1}
	X_j
	=P_j\Sigma_j Q^T_j,
\end{equation}
where $P_j\in \mathbb{R}^{n_j\times r_j}, Q_j\in \mathbb{R}^{m\times r_j}$. Then we have $X_jX^T_j= P_j \Sigma^2_jP^T_j(j=1,\cdots,J)$ and convert $X_j X^T_j W_j= X_jG^T$ into
$$P^T_j W_j = \Sigma^{-1}_{j}Q^T_j G^T, ~~ \forall j=1,\cdots,J,$$
which could be further written as
$$A_jW_j+B_jZ=0, ~~ \forall j=1,\cdots,J,$$
with notation $A_j=P_j^T$, $B_j=-\Sigma^{-1}_{j}Q^T_j$ and $Z=G^T$. With these optimality conditions, we now formulate the sparse solution of the canonical variates $W_j$ as follows,
\begin{align}\label{ell1gcca}
	\min &~~ \|W\|_1=\sum^J_{j=1}\|W_j\|_1\nonumber\\
	\mathrm{s.t.} &~~ A_jW_j+B_jZ=0,\qquad Z^TZ= I_{\ell},
\end{align}
where the $\ell_1$ norm of a matrix is defined as the summation of the $\ell_1$ norm of its columns, i.e.,  $\|W\|_1=\sum^J_{j=1}\|W_j\|_1$. Obviously, unlike Eq. (\ref{maxvar}),  Eq. (\ref{ell1gcca}) imposes sparsity constraints to interpret canonical variables. On the other hand,  from the Bayesian inference viewpoint,  Eq. (\ref{ell1gcca}) could be considered as a MAP estimate of
$\|W\|_1$ with Laplace prior under special constraints. Due to the constraint $Z^TZ=I_\ell$, problem (\ref{ell1gcca}) is a nonconvex problem on Stiefel manifold. Before addressing the convergence analysis, we present the following first-order optimality conditions.
\begin{lemma}\label{mainlem}
Denote $W^*=(W^*_1,W^*_2,\cdots, W^*_J)$, let $(W^*,Z^*)$ be a local minimizer of problem (\ref{ell1gcca}), for each fixed $j=1,\cdots, J$, then there exist Lagrange multipliers $\Lambda^*_{1,j}\in R^{r_j\times \ell},  \Lambda_2^*\in R^{\ell\times \ell}$, $(j=1,\cdots,J)$ such that
\begin{eqnarray}\label{KKTcon}
		&&A^T_j \Lambda^*_{1,j}\in \partial \|W^*_j\|_1,~~\sum^J_{j=1}B^T_j \Lambda^*_{1,j}+Z^* \Lambda^*_2=0 \nonumber\\
		&& A_j W^*_j+B_j Z^*=0,~~ (Z^*)^T Z^*=I_{\ell},
\end{eqnarray}
where $\partial \|W^*\|_1$ stands for the subdifferential of $\|\cdot\|_1$ at $W^*$.	
\end{lemma}
{\bf Remark:} The conditions stated in Eq. (\ref{KKTcon}) are actually KKT conditions achieved by utilizing Lagrange multiplier method, and computing the partial derivatives with respect to $W_j$ $(j=1,\cdots, J)$, $Z$, $\Lambda_{1,j}$ $(j=1,\cdots, J)$ and $\Lambda_2$.
	
\subsection{Sparse GCCA algorithm via distributed ADMM}
To solve the proposed sparse GCCA (\ref{ell1gcca}), we in this subsection will develop an efficient algorithm, mainly in the framework of distributed ADMM. The augmented Lagrangian of (\ref{ell1gcca}) is
\begin{align*}
& L_\beta(W_1,\cdots, W_J,Z,\Lambda_1, \cdots, \Lambda_J)=\sum^J_{j=1} \|W_j\|_1+I_{O_\ell}(Z)\\
&~~~~ -\sum^J_{j=1} \langle \Lambda_j, A_j W_j+B_jZ\rangle +\frac{\beta}{2}\sum^J_{j=1}\|A_jW_j+B_jZ\|^2_F,
\end{align*}
where $O_\ell=\{Z^TZ=I\}$, $\Lambda_j \in R^{r_j\times\ell}$ ($j=1,\cdots,J$) are the Lagrange multipliers corresponding to the constraints $A_jW_j+B_jZ$$ (j=1,\cdots,J)$ . Here we reduce notational burden and remove the subscript $1$ for $\Lambda_{1,j}$,  $I_{O_\ell}(Z)$ is an indicator function defined as
\begin{equation*}
	I_{O_\ell}(Z)=\left\{
	\begin{array}{ll}
	0, & Z\in O_\ell, \\
	+\infty, & \mathrm{otherwise}.
	\end{array}
	\right.
\end{equation*}
Directly applying the classical iteration process of ADMM, we will have
\begin{equation}\label{sgcca_two}
	\left\{
	\begin{array}{ll}
	(W^{k+1}_1,\cdots,W^{k+1}_J,Z^{k+1})=\argmin _{W_1,\cdots, W_J,Z}  &\\
	~~~~~~~~ L_{\beta_k}(W_1,\cdots, W_J,Z,\Lambda_1, \cdots, \Lambda_J), &  \\
	\Lambda^{k+1}_j =\Lambda^k_j-\beta_k (A_j W^{k+1}_j+B_j Z^{k+1}). &
	\end{array}
	\right.
\end{equation}
However, it is difficult to obtain  $W^{k+1}_j$ ($j=1,\cdots,J$) and $Z^{k+1}$ simultaneously. Moreover, the existence of $A_j$ and $B_j$ with different ranks makes it more challenge and the orthogonality constraint $Z^TZ=I$ leads the problem non-convex. Generally, classical ADMM does not work and we need to employ the iteration idea stated in distributed ADMM with slight modifications.
	
First, we decouple the update of $W^{k+1}_j$ ($j=1,\cdots,J$) and $Z^{k+1}$, i.e., $L_{\beta_k}$ is optimized with the respect to variables $W_j$ and $Z$ one at a time, while fixing the others at their latest values. Mathematically, the above idea of updating the Lagrange multipliers could be written as,
\begin{align}
	Z^{k+1} &= \argmin_{Z\in O_{\ell}} L_{\beta_k}(W^k_1,W^k_2,\cdots, W^k_J,Z,\Lambda^k_1,\Lambda^k_2,\cdots,\Lambda^k_J),\label{iterZ}\\
	W^{k+1}_j &= \argmin_{W_j} L_{\beta_k}(W_j,Z^{k+1}, \Lambda^k_1,\Lambda^k_2,\cdots, \Lambda^k_J),\label{iterW}\\
	\Lambda^{k+1}_j &= \Lambda^k_j-\beta_k(A_jW^{k+1}_j+B_j Z^{k+1}),
\end{align}
where $L_{\beta}(W_j,Z$, $\Lambda_1$,  $\Lambda_2$, $\cdots$, $\Lambda_J)$ stands for  $L_\beta(W_1$, $\cdots$, $W_J$, $Z$, $\Lambda_1$, $\cdots$, $\Lambda_J)$ with fixed $W_i(i\neq j)$.
In fact, Eq. (\ref{iterZ}) could be simplified as
\begin{align*}
Z^{k+1} &= \argmin_{Z\in O_{\ell}} \Big\{-\sum^J_{j=1}\langle \Lambda^k_j, A_j W^k_j+B_j Z\rangle\\
&+\frac{\beta_k}{2}\sum^J_{j=1} \|A_j W^k_j+B_jZ\|^2_F\Big\},\\
&= \argmin_{Z\in O_{\ell}} \Big\{\sum^J_{j=1}\|A_j W^k_j+B_jZ-\frac{ \Lambda^k_j}{\beta_k} \|^2_F\Big\}.\\
\end{align*}
Recall the global consensus problem \cite{Boyd2011},
\begin{align}\label{globalcon}
	\min &~~ \sum^N_{i=1}f_i(x_i)\nonumber\\
	\mathrm{s.t.} &~~x_i-z=0,~~i=1,\cdots,N,
\end{align}
which described an optimization problem under the constraints that all the local variables should agree. In fact, the constraints in Eq. (\ref{ell1gcca}) could be viewed as a weighted matrix version of those in Eq. (\ref{globalcon}). $Z$ plays the role of a central collector. Therefore, motivated by consensus ADMM method discussed in \cite{Boyd2011}, we propose the distributed alternating iteration based sparse gcca algorithm. We can achieve a new iteration formula for $Z$ as indicated in Eq. (\ref{admmgcca1}). Since $W^k_1$, $\cdots$, $W^k_i(i\neq j)$, $W^k_J$ are fixed, and notice the definition of $L_{\beta_k}(W_j,Z$, $\Lambda_1$,$\Lambda_2$,$\cdots$, $\Lambda_J)$,  we can further simplify Eq. (\ref{iterW}) as
 \begin{align*}
W^{k+1}_j &= \argmin_{W_j} \Big\{\|W_j\|_1-\langle \Lambda^k_j, A_j W_j+B_j Z^{k+1}\rangle\\
&+\frac{\beta_k}{2}\|A_j W_j+B_jZ^{k+1}\|^2_F\Big\}.
\end{align*}
Therefore, simple computation leads to
\begin{equation}\label{admmgcca1}
	Z^{k+1} = \argmin_{Z\in O_\ell} \Big\{\|\bar W^k+\bar B Z-\frac{1}{\beta_k} \bar \Lambda^k\|^2_F \Big\}
\end{equation}
\begin{equation}\label{admmgcca2}
	W^{k+1}_j = \argmin\Big\{\|W_j\|_1+\frac{\beta_k}{2}\|A_j W_j+B_j Z^{k+1}-\frac{\Lambda^k_j}{\beta_k}\|^2_F\Big\}
\end{equation}
\begin{equation}\label{admmgcca3}
	\Lambda^{k+1}_j = \Lambda^k_j-\beta_k(A_jW^{k+1}_j+B_j Z^{k+1}).
\end{equation}
The key point is to effectively solve problem (\ref{admmgcca1}), which embody the central collector role of $Z$ and to achieve the analytic expression for $\bar W^k$, $\bar\Lambda^k$, and $\bar B$.  For different $i$, the number of rows of $A_i$, $B_i$, and $\Lambda_i$ may be different. For the sake of clarity,  we consider the augmented version of $A_j$, $B_j$ and $\Lambda_j$ ($j=1,\cdots, J$), i.e., rows with zero entries are added, such that  $A_j$, $B_j$, and $\Lambda_j$ all have $r$ rows with $r=\max\{r_j, j=1,\cdots,J\}$. Let ${\tilde A}_j, {\tilde \Lambda}_j, {\tilde B}_j$ be the augmented version of $A_j$, $\Lambda_j$ and $B_j$, respectively. Then $\bar W^k$, $\bar\Lambda^k$ and $\bar B$ can be calculated as follows,
$$\bar W^k=\frac1J\sum^J_{j=1} {\tilde A}_jW^k_j, \quad \bar\Lambda^k=\frac1J\sum^J_{j=1} {\tilde \Lambda}^k_j, \quad\bar B=\frac1J\sum^J_{j=1}{\tilde B}_j.$$
For  Eq. (\ref{admmgcca1}), we find that this optimization task is actually a Procrustes problem \cite{golub1996matrix} and could be solved
analytically as the following,
\begin{equation}\label{updateZ}
	Z^{k+1} = U^k (V^k)^T,
\end{equation}
where $U^k,V^k$ are orthogonal matrices from the SVD of the matrix
$${\bar B}^T \left(\frac{{\bar \Lambda}^k}{\beta_k}-{\bar W}^k\right) = U^k \Sigma^k (V^k)^T.$$
For Eq. (\ref{admmgcca2}), this optimization task is an $\ell_1$-norm regularized least squares problem, which does not have a closed-form
solution. To avoid an exhaustive iterative process, we approximate it by linearizing the Frobenius norm term (for each fixed $j$) as below,
\begin{align}\label{iteration2}
	&\min_{W_j} ~ \|W_j\|_1+ \beta_k \Big\{ \langle A^T_j(A_j W_j^k+B_j Z^{k+1}-\Lambda^k_j/\beta_k),\nonumber \\
	& ~~~~~~~ W_j-W^k_j\rangle +\|W_j-W^k_j\|^2_F/2\delta\Big\},
\end{align}
where $\delta>0$ is a proximity parameter. Then  Eq.  (\ref{admmgcca2}) can be approximately solved as
\begin{equation}\label{updateW}
	W^{k+1}_j =S\Big(W^k_j-\delta A^T_j (A_jW^k_j+B_j Z^{k+1}-{\Lambda^k_j}/{\beta_k}), \frac{\delta}{\beta_k}\Big),
\end{equation}
where $S(x,\mu)$ is  the componentwise soft-thresholding shrinkage operator defined as
$$S(x,\mu)={\rm sgn}(x)\odot \max\{|x|-\mu,0\}$$
with $\odot$ denoting the componentwise products of vectors or matrices. This update gives an approximate but closed-form solution for Eq.  (\ref{admmgcca2}).
	
Now, we come to the following augmented distributed alternative iteration based sparse GCCA algorithm, as summarized in Algorithm \ref{SGCCA}.
\begin{algorithm}[!ht]
		\renewcommand{\algorithmicrequire}{\textbf{Input:}}
		\renewcommand\algorithmicensure {\textbf{Output:} }
		\caption{Distributed alternative iteration based sparse generalized CCA algorithm (SGCCA).}
		\label{SGCCA}
		\begin{algorithmic}[1]
			\REQUIRE ~~\\
			Training data  $X_j\in \mathbb{R}^{n_j\times m}$ ($j=1,\cdots,J$), parameter $\delta>0$, $\rho>1$, $\beta_{\max}$ and tolerance parameter $\varepsilon_1, \varepsilon_2$.\\
			\ENSURE ~~\\
			Sparse canonical variates $W=(W^T_1,\cdots,W_J^T)^T$.\\
			\STATE Compute reduced SVD for each $X_j$ ($j=1,\cdots,J$) via Eq. (\ref{decomp1}).
			
			\STATE Let $W_j^0=\Lambda_j^0=0$, $\beta_0=\max(1/\|A^T_jB_j\|_\infty)$,  $(j=1,\cdots, J)$.
			\WHILE {$\|\Lambda^{k+1}_j-\Lambda^k_j\|_F/\beta_k>\varepsilon_1$ and $\beta_k\|W^{k+1}_j-W^k_j\|_F/\max\{1,\|W^k_j\|_F\}>\varepsilon_2$}
			\STATE {Compute $Z^{k+1}$, $W^{k+1}_j$ and $\Lambda_j^{k+1}$  via Eqs. (\ref{updateZ}), (\ref{updateW}) and (\ref{admmgcca3}), respectively.}
			\STATE  {Update $\beta_{k+1}$ by $\beta_{k+1}=\min (\beta_{\max}, \rho\beta_k)$}
			\ENDWHILE
		\end{algorithmic}
\end{algorithm}
	
In Algorithm \ref{SGCCA}, all the updates have analytical expressions and work very efficiently. But its convergence can not be naturally inherited from classical ADMM, since the original problem (\ref{ell1gcca}) is non-convex and the update for Eq. (\ref{admmgcca2}) is inexact. Based upon Lemma \ref{mainlem}, we can now give the theoretical convergence analysis.

\begin{theorem}\label{mainthm}
Assume that $A_j$ is of full row rank for each fixed $j$$(j=1$,$\cdots$, $J)$. Let $(W^k_1,\cdots, W^k_J, Z^k,\Lambda^k_1,\cdots, \Lambda^k_J)$ be generated by solving subproblems $(\ref{admmgcca1})-(\ref{admmgcca3})$ exactly and $\beta_{k+1}=\rho \beta_k(\rho>1)$. Then, the sequence  $(W^k_1,\cdots, W^k_J, Z^k,\Lambda^k_1,\cdots, \Lambda^k_J)$ is bounded, and
$$\lim_{k\to\infty} A_j W^k_j+B_j Z^k=0 ~(j=1,\cdots, J).$$
Moreover, any accumulation point $(W^*_1$,$\cdots$, $W^*_J$,$Z^*$, $\Lambda^*_1$, $\cdots$, $\Lambda^*_J$, $-\sum^J_{j=1}(B_jZ^*)^T\Lambda^*_j)$ of $\{(W^k_1$,$\cdots$, $W^k_J$, $Z^k$, $\Lambda^k_1$,$\cdots$, $\Lambda^k_J$,$-\sum^J_{j=1}(B_jZ^{k})^T\Lambda^k_j)\}^\infty_{k=1}$ satisfies the KKT conditions (\ref{KKTcon}). In particular, whenever   $\{(W^k_1$,$\cdots$, $W^k_J$, $Z^k$, $\Lambda^k_1$,$\cdots$, $\Lambda^k_J$,$-\sum^J_{j=1}(B_jZ^k)^T\Lambda^k_j)\}^\infty_{k=1}$  converges, it converges to a KKT point of problem  (\ref{ell1gcca}).
\end{theorem}

Under mild conditions, any limit point of the iterative sequence generated by Algorithm \ref{SGCCA} is a KKT  point of problem (\ref{ell1gcca}). Based upon the convergence conditions of matrix series, we  remove the ``F" symbol and  get the following convergence analysis for Algorithm \ref{SGCCA}.
\begin{theorem}\label{mainthm2}
		Let	$\{(W^k_1$,$\cdots$, $W^k_J$, $Z^k$, $\Lambda^k_1$,$\cdots$, $\Lambda^k_J)$$\}^\infty_{k=1}$ be the sequence generated by Algorithm \ref{SGCCA}. For each fixed $j (j=1,\cdots,J)$, assume that $\lim_{k\to\infty} \beta_k(W^{k+1}_j-W^k_j)=0$ and $\lim_{k\to\infty}(\Lambda^{k+1}_j-\Lambda^k_j)/\beta_k=0$. Then any accumulation point $(W^*_1$,$\cdots$, $W^*_J$,$Z^*$, $\Lambda^*_1$, $\cdots$, $\Lambda^*_J$, $-\sum^J_{j=1}(B_jZ^*)^T\Lambda^*_j)$  of $\{(W^k_1$,$\cdots$, $W^k_J$, $Z^k$, $\Lambda^k_1$,$\cdots$, $\Lambda^k_J$,$-\sum^J_{j=1}(B_jZ^k)^T\Lambda^k_j)\}^\infty_{k=1}$  satisfies the KKT conditions (\ref{KKTcon}). In particular,  whenever $\{(W^k_1$,$\cdots$, $W^k_J$, $Z^k$, $\Lambda^k_1$,$\cdots$, $\Lambda^k_J$,$-\sum^J_{j=1}(B_jZ^k)^T\Lambda^k_j)\}^\infty_{k=1}$  converges, it converges to a KKT point of problem (\ref{ell1gcca}).
\end{theorem}

The proofs of Lemma \ref{mainlem}, Theorems \ref{mainthm} and \ref{mainthm2} are given in the apendix.

\subsection{Sparse GCCA with fixed $G$}
If $G$ is fixed, we can get a much simpler version for the proposed sparse GCCA. Denote $W_j=(\alpha^1_j,\cdots,\alpha^\ell_j)$, $Z=(z_1,\cdots,z_\ell)$, where  the columns of $Z=G^T$ are the eigenvectors of the matrix
$$M=\sum_{j=1}^J X^T_j (X_jX^T_j)^{-1}X_j.$$
For each fixed $i$, we need to solve $J$ problems:
$$P^T_j\alpha^i_j=\Sigma^{-1}_jQ^T_jz_i,~~(i=1,\cdots,\ell).$$
To achieve sparsity of canonical variates, we establish the following model,
\begin{align}\label{fistagcca}
	&{\rm min}~~ \|\alpha^i_j\|_1\nonumber\\
	&s.t.~~ P^T_j \alpha^i_j=\Sigma_j^{-1}Q_j^Tz_i, ~j=1,\cdots,J.
\end{align}
To reduce notational burden and allow a slightly abuse of notation, we omit the index $i,j$ in the sequel unless specified. As a classical $\ell_1$ problem, there are lots of effective algorithms for Eq.  (\ref{fistagcca}), such as Iterative Shrinkage-Thresholding Algorithm (ISTA, \cite{Wright2008SparseRB}, \cite{Hale2007TR0A}), Least Angle Regression (LARS), subgradient descent. Here we utilize fast ISTA (FISTA, \cite{Beck2009AFI}).
Let $v_1=\alpha_0, t_1=1$. The update for the sparse GCCA with  fixed $G$ is given below,
\begin{equation}\label{sgcca_fixedG}
	\left\{
	\begin{array}{ll}
	\alpha_s=\argmin_{\alpha\in {\mathbb R}^n} \Big\{\|\alpha\|_1+ \frac{L}{2}\Big\|\alpha-(v_s- &\\
	~~~~~~~~~~ \frac{2}{L}(PP^Tv_s-P\Sigma^{-1}Q^Tz))\Big\|^2_2\Big\}, &\\
	t_{s+1}=\frac{1+\sqrt{1+4t_s^2}}{2}, &\\
	v_{s+1}=\alpha_s+\frac{t_s-1}{t_{s+1}}(\alpha_s-\alpha_{s-1}), &
	\end{array}
	\right.
\end{equation}
where $L=\|PP^T\|$ differs for a distinct view and  $t_s$ is the stepsize of iteration.
\section{Experimental results}\label{section4}
In this section, we carry out numerical experiments on both synthetic dataset and real-world datasets to evaluate the proposed sparse GCCA algorithm by comparing it with other algorithms.
All the experiments are performed under Ubuntu 16.04 with python 3.7 in Intel(R) Xeon(R) CPU E5-2620 v4 @ 2.10GHz and 128 GB of RAM.

\subsection{Experiments on Synthetic dataset}
First, we consider the proposed sparse GCCA (SGCCA) on synthetic data to evaluate its convergence, sparsity, and accuracy. Three matrices $X$ , $Y$, and $Z$, are constructed as the following,
$$X=v_1u^T+\epsilon_1,\qquad Y=v_2u^T+\epsilon_2,\qquad  Z= v_3u^T+\epsilon_3,$$
where
$$v_1=(\underbrace{1,\cdots,1}_{2000}, \underbrace{-1,\cdots, -1}_{3000}, \underbrace{0,\cdots, 0}_{5000})^T,$$
$$v_2=(\underbrace{0,\cdots,0}_{10000}, \underbrace{1,\cdots,1}_{2000},\underbrace{-1,\cdots,-1}_{3000})^T,$$
$$v_3=(\underbrace{1,\cdots,1}_{2000}, \underbrace{0,\cdots,0}_{12000}, \underbrace{-1,\cdots,-1}_{3000})^T,$$
and $\epsilon_1\in \mathbb{R}^{10000\times 100}$ $\sim {\cal N}(0,0.3^2)$, $\epsilon_2\in \mathbb{R}^{15000\times 100}$ $\sim {\cal N}(0,0.4^2)$,  $\epsilon_3\in \mathbb{R}^{17000\times 100}\sim {\cal N}(0,0.5^2)$ are three random noise matrices.
$u\in R^{100}$ is a random vector with all entries drawn from the normal distribution, i.e.,
$$u(i)\sim {\cal N}(0,1), ~~i=1,\cdots, 100.$$

{\bf Settings}~~We randomly select half of the data for training and use the rest for test. This procedure is repeated $30$ times and the average results are reported. GCCA is used as the baseline algorithm. We also consider  weighted GCCA (WGCCA, \cite{benton2016learning}) with Gaussian initialization weights and DGCCA (\cite{benton}) with $100$ ${\rm epochs}$ and batch size tuned by cross-validation. For the parameters of the proposed sparse GCCA algorithm, we set $\beta_{\max} = 10^4$, tolerance $\varepsilon_1 =  \varepsilon_2 = 10^{-5}$, and $\delta, \rho$ are tuned by cross-validation around one.

\begin{figure}[!htbp]
	\centering	
	\includegraphics[scale=0.35]{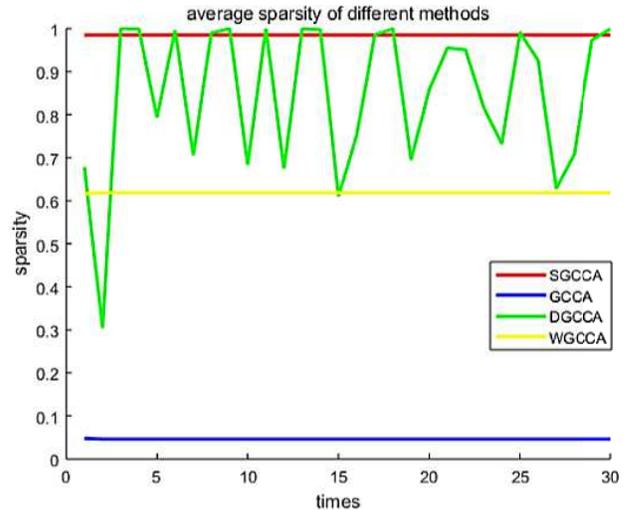}	
	\caption{Average sparsity results over 30 experiments.}
	\label{30times}	
\end{figure}

{\bf Results}~~The aim of this experiments is to find a sparse decomposition structure, in which we could achieve  high accuracy, high sparsity and thus good signal recovery  performance. For all the considered algorithms, the reconstruction error is blow $0.01$. The sparsity is shown by Figure \ref{30times}, which depicts the results of average sparsity over $30$ trials. Here, sparsity denotes the percentage of zero entries of a vector. The proposed GCCA algorithm has very prominent sparsity in the three views and the performance is quite stable.

\subsection{Experiments on Real datasets}
Via experiments on synthetic data, we verify that the proposed sparse GCCA could indeed find the sparse structures.  Here we present the experimental results on gene expression data.
\subsubsection{Gene Expression Data}~~

Let us consider four real datasets from the gene expression database \footnote{ \url{http://stat.ethz.ch/~dettling/bagboost.html}}. The details are explained below and the statistics can be found in Table \ref{feasta}. In this case, we apply the proposed method to the application of classification problem. There are two views, one view is the data matrix $X_1$, another view is the label $X_2$.
\begin{itemize}
	\item Leukemia: gene expression values  for $72$ samples ($47$ samples from patients with acute lymphoblastic leukemia, and $25$ from patients with acute myeoblastic leukemia).
	
	\item Prostate: gene expression values measured by Affymetrix human $95Av2$ arrays for $102$ samples ($52$ prostate tumors and $50$ nontumors prostate samples).
	
	\item Brain: $42$ microarray gene expression profiles from five different tumors of the central nervous system.	

   \item Lymphoma: $42$ samples of diffuse large B-cell lymphoma, $9$ observations of follicular lymphoma and $11$ cases of chronic
	lymphocytic leukemia with the expression of $4026$ well-measured genes. The total sample size is $62$.
\end{itemize}

\begin{table}[!htb]
	\begin{center}
	\caption{Data structures: data dimension ($n$), number of data ($m$), number of classes ($K$), number of columns in $W_1$ and $W_2$ ($\ell$).}\label{feasta}		
		\begin{tabular}{c|ccccc}
			\hline
			Type & Data & $n$   &    $m$    & $K$    &  $\ell$\\
			\hline
			\multirow{5}{*}{Gene Data}
			& Leukemia & 3571    &   72   &   2  &  1 \\

			&Prostate &   6033    &  102   &  2  &1\\
			
			& Brain& 5597    &    42   &    5    &  4 \\

           &Lymphomia & 4026   &  62   & 3   & 2\\	

			\hline
		\end{tabular}
	\end{center}
\end{table}

{\bf Settings}~~The preprocessing procedure for gene data is described in \cite{dettling2004bagboosting}.
In the experiment, we choose $\ell$ as the rank of the matrix $X_1X_2^T$. We compare the proposed sparse GCCA algorithm with GCCA, DGCCA and WGCCA methods by considering several indices: the correlations for training and testing, reconstruction error for training, sparsity of canonical variates  and the classification accuracy for training and testing. The correlation is defined as
$$\sum_{i\neq j}  {\rm Trace}(W^T_i X_iX^T_j W_j),$$
and the reconstruction error is computed as
$$\frac{1}{\ell}\sum^J_{j=1}\|W^T_j X_j-G\|^2_F,$$
while the classification accuracy is defined as the following:
$$\mbox{Accuracy:} =\frac{\sum^m_{i=1}\delta (ol_i,pl_i)}{m},$$
where $\delta(y_1,y_2)$ is the indicator function that equals $1$ if $y_1=y_2$ and $0$ otherwise. For a given sample point $x_i$, $ol_i$ and $pl_i$ are the obtained label and the provided label, respectively.
Among those indices,  sparsity and classification accuracy are the most important.  We select $4/5$ of the data for training and use the rest for test.  The choices of the other parameters are  the same  as that described in the last section.
\begin{table*}[htb]
	\begin{center}
		\caption{Comparison results on  gene expression data.}
\label{gene_result}		
		\begin{tabular}{c|ccccc}
			\hline
			Data &Criterion & SGCCA  &  GCCA    & DGCCA   &  WGCCA\\
			\hline
			\multirow{5}{*}{Leukemia}
			& \bf a1 & 0.9991     &  \bf 1.0000 &   0.4115  &  0.9773 \\
			
			&\bf a2& \bf 0.9203    &  0.9143   & 0.6398   & 0.8971\\

			&\bf a3 & 7.53E-04	  &\bf 6.97E-17	&0.0297&	0.0037\\

			&\bf a4& 0.9826   &0.04621&\bf 	0.9936	&0.0484
			\\
			&\bf a5&\bf  1.0000&  	\bf 1.0000&	0.6316&\bf	1.0000
			\\
			&\bf a6& \bf 1.0000&  \bf1.0000&	0.7333&\bf	1.0000
			\\
			\hline
			\multirow{5}{*}{Prostate}
			& \bf a1 & 0.9995 &\bf 	1.0000&	0.0024&	0.9858
			\\
			
			&\bf a2&\bf  0.8475  &	0.7741&	0.5046&	0.7354
			\\
			
			&\bf a3 & 3.85E-04  &	\bf 7.70E-17&	0.1537&	0.0021
			\\

			&\bf a4& \bf 0.9862 &	0.0560&	0.7086&	0.0612
			
			\\
			&\bf a5&\bf  1.0000 &	\bf1.0000&	0.6049&	\bf1.0000
			\\
			&\bf a6&\bf  0.9048     &	0.9048&	0.4286&	\bf 0.9048
			\\
			\hline
			\multirow{5}{*}{Brain
			}
			& \bf a1 & \bf 1.1593&   	1.0067&	0.0570&	0.9860
			
			\\
			
			&\bf a2& 1.1237    &\bf 	1.1911	&0.3932&	0.7015
			
			\\
			
			&\bf a3 & 0.1033 &\bf 	0.0909&	0.4649&	0.0963
			
			\\

			&\bf a4& 0.9462&  0.0400&	\bf 0.9951&	0.0686

			\\
			&\bf a5&\bf  1.0000&  \bf1.0000&	0.1515&	0.9394
			\\
			&\bf a6& \bf 0.4444&    0.2222&	0.1111&	0.2222
			\\
			\hline
			\multirow{5}{*}{Lymphomia
			}
			& \bf a1 &\bf 1.3041 &	1.0069	&1.0825&	0.9987
			
			\\
			
			&\bf a2&\bf  1.3876&	1.3459	&1.3662	&1.2618
			
			\\
			
			&\bf a3 & 0.0208&\bf 	0.0204&	0.0614&	0.0229
			
			\\

			&\bf a4& 0.9839&  0.0396&	\bf 0.9898	&0.0606

			\\
			&\bf a5&\bf  1.0000&  \bf 1.0000&	0.3061&	\bf 1.0000
			\\
			&\bf a6& \bf 1.0000& \bf 1.0000&	0.5385&	\bf 1.0000
			\\
			\hline
		\end{tabular}	
	\end{center}
\end{table*}

{\bf Results}~~The detailed comparison results are shown in Table \ref{gene_result}, which contains the following criteria.
\begin{itemize}
	\item $\bf a1$ denotes the correlation for training data.
	\item $\bf a2$ means the correlation for test data.
	\item $\bf a3$ stands for the reconstruction error for training.
	\item $\bf a4$ denotes the sparsity of canonical variate (related to the data matrix).
	\item $\bf a5$ means the classification accuracy using sparse $W$ in training data.
	\item $\bf a6$ stands for the classification accuracy using sparse $W$ in test data.
\end{itemize}

The best results are marked in bold. The proposed  sparse GCCA algorithm is competitive with the other GCCA methods in accuracy related criterion and it performs remarkably in sparsity without much loss in classification accuracy.

\subsubsection{Cross-Language Document Retrieval}~~
In this section, we conduct experiments on Europarl parallel corpus (\cite{Koehn_europarl:a}, Europarl for short), which is a collection of documents extracted from the proceedings of the European Parliament. It includes translated documents in $21$ European languages: Romanic (French, Italian, Spanish, Portuguese, Romanian), Germanic (English, Dutch, German, Danish, Swedish), Slavik (Bulgarian, Czech, Polish, Slovak, Slovene), Finni-Ugric (Finnish, Hungarian, Estonian), Baltic (Latvian, Lithuanian), and Greek. The main aim is to learn the latent representations of the sentences, which reveals the correlations of  the same sentences in different languages (views).

{\bf Settings}~~
For Europarl dataset, we select three different types of language data (English, French, and Spanish), and obtain a bag-of-words representation using Term Frequency Inverse Document Frequency (TFIDF) approach, which is widely recognized as an efficient way in the document retrieval task. After removing numbers, stop-words (English, French, and Spanish, respectively), and rare words (appearing less than twice), we obtain distinct sizes of matrices for different type of language data, which is demonstrated in  Table \ref{fealang}. We compare the proposed sparse GCCA algorithm with GCCA, DGCCA and WGCCA methods for $\ell=1$ by considering several indices: the correlations for training and test, reconstruction error for training, sparsity of canonical variates  and average area under the ROC curve (AROC, \cite{Sriperumbudur2011A}) for both training and test. We compute the Euclidean distance between every projected view and latent lower dimensional space, and sort them in an increasing order to match the most relevant document. We choose half of the data as training and use the rest for test. The parameters for the other methods are the same as that described in the previous section.
\begin{table}[!htb]
	\begin{center}
	\caption{Data structures: data dimension ($n$), number of data ($m$), number of columns in $W_1$, $W_2$ and $W_3$ ($\ell$).}\label{fealang}		
\begin{tabular}{c|ccccc}
			\hline
			Type & Data & $n$   &    $m$     &  $\ell$\\
			\hline
			\multirow{4}{*}{Document Data}
			&English & 5552    &   1000   &  1 \\
			
			&French & 6126   &  1000      & 1\\	
			
			&Spanish &   6292    &  1000    &1\\
			
			\hline
		\end{tabular}
	\end{center}
\end{table}

{\bf Results}~~The detailed performance compared with other algorithms is reported in Table \ref{language_result1}. The AROC results are shown in Table \ref{language_result2}. We evaluate AROC for each pair of languages in Table \ref{language_result2}, i.e.,   \RNum{1} denotes the retrieval results of English-French pair, \RNum{2} means  the retrieval results of English-Spanish pair, \RNum{3} stands for the retrieval results of  French-Spanish pair and $\rm null$ denotes average AROC results of the above three. These tables include the following criteria.
\begin{itemize}
	\item $\bf b1$ denotes the correlation for training data.
	\item $\bf b2$ means the correlation for test data.
	\item $\bf b3$ denotes the reconstruction error for training.
	\item $\bf b4$ stands for the sparsity of canonical variates for each view.
	\item $\bf b5$ stands for the average sparsity of canonical variates.
	\item $\bf b6$ denotes the AROC results for training data.
	\item $\bf b7$ means the  AROC results of each pair for test data.
	\item $\bf b8$ denotes the average AROC results for test data.
\end{itemize}

\begin{table}[htb]
	\begin{center}
	\caption{Comparison results on Europarl dataset.}\label{language_result1}
		\begin{tabular}{c|ccccc}
			\hline
			Criterion & SGCCA  &   GCCA    &  DGCCA   & WGCCA\\
			\hline
			$\bf b1$
			& 0.9971  &   \bf 1.0000&       0.0283  & 1.0000\\
			\hline
			$\bf b2$
			& 0.0244 &\bf 0.4030    &0.0401 &0.0228    \\
			\hline
			$\bf b3$&
			1.49E-03&\bf 1.31E-30       &7.23E+03  &2.45E-11\\
			\hline
			\multirow{4}{*}{$\bf b4$}
			&\bf 0.8559   &0.1059      &0.0904   &0.1081  \\
			&
			\bf 0.8767    &0.1100     &0.9987   &0.1116\\
			&\bf 0.8755   &0.1187     &0.2146  &0.1219  \\
			\hline
			$\bf b5$     &\bf 0.8694   &0.1115  &0.4346 &0.1139\\
			
			\hline
		\end{tabular}	
	\end{center}
\end{table}
Obviously, the proposed sparse GCCA algorithm is considerably more prominent than other algorithms in sparsity for every view and also slightly outperforms in average AROC, showing that SGCCA could preferably find the sparse structure in the cross-language document retrieval task.
\begin{table}[htb]\small
	\begin{center}
	\caption{AROC achieved by Europarl dataset.}
     \label{language_result2}		
		\begin{tabular}{c|c|cccc}
			\hline
			Criterion &type & SGCCA  &   GCCA    & DGCCA   &  WGCCA\\
			
			\hline
			\multirow{5}{*}{\shortstack{$\bf b6$}}
			&\RNum{1}&  0.8325 &  \bf 1.0000  &  \bf 1.0000  &0.4956\\
			
			&\RNum{2}&  0.6905  & \bf 1.0000   & \bf 1.0000     &0.4867 \\
			&\RNum{3}&   0.7072 & \bf 1.0000   & \bf 1.0000    & 0.5008\\
			
			\hline
			\multirow{5}{*}{\shortstack{$\bf b7$}}
			&\RNum{1}&  \bf 0.6198 &  0.5773  &   0.5173  &0.4600\\
			
			&\RNum{2}&  0.5837  &\bf 0.5840   &0.5239    &0.5155\\
			&\RNum{3}&   0.5823 & \bf 0.6111  & 0.5112   & 0.5009\\
			\hline
			\shortstack{$\bf b8$} &${\rm null}$&\bf 0.5953&0.5908&0.5175&0.4921\\
			
			\hline
		\end{tabular}	
	\end{center}
\end{table}

\section{Conclusion}\label{conclu}
In this paper, based on MAX-VAR formulation of GCCA, by employing SVD technique, we achieved a novel GCCA framework from the linear system of equations viewpoint, and imposed sparsity under this framework. Theoretical consitency was investigated under mild condtitions. We designed a distributed alternating iteration based sparse GCCA  algorithm. Experimental results  on synthetic dataset, gene data, and Europarl dataset all demonstrated the effectiveness of the proposed algorithm, which is promising for CCA applications that involves more than two views and have sparsity structure.

	\section*{Appendix}\label{section:appe}
	This section gives the proofs of Lemma 1, Theorems 1 and 2. Before that, we directly cite two lemmas which play significant role in proving  theoretical results. We first give some notations.

The effective domain of a convex function $f$ on $S$ is denoted by ${\rm dom}f$, and defined by
$${\rm dom}f=\{x|\exists \mu, (x,\mu)\in {\rm epi} f\}=\{x|f(x)< +\infty\},$$
where ${\rm epi} f$ stands for the epigraph of $f$. $ri C$ means the relative interior of a convex set C:
$$ri C =\{x\in {\rm aff} C|\exists \varepsilon >0, (x+\varepsilon B)\bigcap ({\rm aff}  C)\subset C\},$$
where ${\rm aff}  C$ means the affine hull of $C$,  $B =\{x|\|x\|\leq 1\}$ denotes the Euclidean ball in  $R^n$. $f'(x;y)$ stands for the directional derivative of $f$ at $x$ with the respect to a vector $y$.
\begin{lemma}(\cite{rockafellar1970convex},Theorem 23.4)\label{lemma2}
		Let $f$ be a proper convex function. For $x\notin {\rm dom} f$, $\partial f(x)$ is empty. For $x\in ri ({\rm dom} f)$, $\partial f(x)$ is non-empty, $f'(x;y)$ is closed and proper as a function of $y$, and
		$$f'(x;y)=\sup \{\langle x^*,y\rangle |x^*\in \partial f(x)\}=\delta^* (y|\partial f(x)).$$
		Finally, $\partial f(x)$ is a non-empty bounded set if and only if $x\in int({\rm dom} f)$  ($int$ stands for the interior of a set), in which case $f' (x;y)$ is finite for every $y$.
\end{lemma}

\begin{lemma}(\cite{rockafellar1970convex}, Theorem 24.4)\label{lemma3}
Let $f$ be a closed proper convex function on $R^n$. If $x_1,x_2,\cdots$, and $x^*_1, x^*_2,\cdots$, are two sequences such that
$x^*_i\in \partial f(x_i)$, where $x_i$ converges to $x$ and $x^*_i$ converges to $x^*$, then $x^*\in \partial f(x)$. In other words, the graph of $\partial f$ is a closed subset of $R^n\times R^n$.		
\end{lemma}

Following the similar technique route for ADMM convergence discussion (\cite{zhang2015sparse}), we can prove Lemma 1 as the following,
\begin{proof}[Proof of Lemma 1]
Since $(W_1^*,\cdots, W^*_J, Z^*)$ is a local minimizer of problem (7), it should satisfy the optimality conditions,
$$A_j W^*_j +B_j Z^*=0,~~(Z^*)^T Z^*=I_{\ell}.$$
By proving the existence of $\Lambda_{1,j}$ $(j=1,\cdots, J)$ such that $A^T_j$ $\Lambda^*_{1,j}\in \partial \|W^*_j\|_1$ for each fixed $j$ and setting $\Lambda^*_2 =-\sum^J_{j=1}(B_jZ^*)^T\Lambda^*_{1,j}$, we have the conclusion that $(W^*_1$, $\cdots$, $W^*_J$, $Z^*$, $\Lambda^*_{1,1}$, $\cdots$, $\Lambda^*_{1,J}$, $\Lambda^*_2)$  satisfies the optimality conditions in (8). For each fixed $j$, to prove the existence of such $\Lambda^*_{1,j}$, we only need to prove $S_j\bigcap \partial{\cal J}(W^*_j)\neq \emptyset$, where
the cone $S_j=\{A^T_j \Lambda_{1,j}, \Lambda_{1,j}\in R^{r_j\times \ell}\}$,  and ${\cal J}(W_j)=\|W_j\|_1$, $j=1,\cdots, J$. Obviously, both $S_j$ and  ${\cal J}(W_j)$ are nonempty closed convex sets. Suppose that   $S_j\bigcap \partial {\cal J}(W^*_j)= \emptyset$. According to the separation theorem of convex sets (\cite{rockafellar1970convex}),  there exist nonzero $Y_j\in \mathbb R^{n_j\times \ell}$ for each fixed $j$ $(j=1,\cdots, J)$ such that		
$$\langle Y_j, D_j\rangle \leq \langle Y_j, A^T_j\Lambda_{1,j}\rangle -1,~\forall D_j\in \partial {\cal J}(W^*_j), ~\Lambda_{1,j}\in R^{r_j\times \ell}.$$
Thus, $A_jY_j=0,j=1,\cdots, J$, otherwise, let $\Lambda_{1,j}= \alpha A_jY_j$ and $\alpha\to -\infty$ so that $\langle Y_j, D_j\rangle \leq -\infty$ which is obviously false. Hence, we can see that
$$\langle Y_j, D_j\rangle \leq -1, \forall D_j\in \partial {\cal J}(W^*_j), ~~j=1,\cdots, J.$$
Let $W_j (\alpha)= W^*_j +\alpha Y_j$ $(j=1,\cdots, J)$, then $(W_1(\alpha), \cdots, W_J(\alpha), Z^*)$ is a feasible point of problem (7) and $W_j(\alpha)\to W^*_j$ as $\alpha\to 0^+$. Since $(W^*_1,\cdots, W^*_J, Z^*)$ is a local minimizer of problem (7), we have
$${\cal J}(W_j(\alpha))-{\cal J}(W^*_j)\geq 0, ~~j=1,\cdots, J$$
for a sufficiently small $\alpha$, and the directional derivative of ${\cal J}$ at $W^*_j$ is defined as
$${\cal J}' (W^*_j, Y_j)= \lim_{\alpha\to 0^+} \frac{{\cal J}(W_j(\alpha))-{\cal J}(W^*_j)}{\alpha}\geq 0.$$
However, Lemma 2 tells us that
$${\cal J}' (W^*_j, Y_j)= \max _{D_j\in \partial {\cal J}(W^*_j)} \langle Y_j, D_j\rangle \leq -1,j=1,\cdots, J.$$
which is a contradiction. Hence, for each fixed $j$, $S_j\bigcap \partial{\cal J}(W^*_j)\neq \emptyset.$
\end{proof}
	
Now we are in position to give the proof of Theorem 1.
\begin{proof}(Proof of Theorem 1)
We only need to prove the boundedness of $\{W^k_j\}^\infty_{k=1}$ and $\{\Lambda^k_j\}$ for each $j=1,\cdots, J$, since $\{Z^k\}$ is an orthogonal matrix which  is obviously bounded.
Because  $W^{k+1}_j$ solves (14), it satisfies the optimality condition
$$0\in \beta_k A^T_j(A_j W^{k+1}_j+B_j Z^{k+1}-\Lambda^k_j/\beta_k)+\partial \|W^{k+1}_j\|_1, ~~\forall k\geq 0,$$
or equivalently
$$A^T_j \Lambda^{k+1}_j\in \partial\|W^{k+1}_j\|_1,$$
by noticing  $\beta_k(A_j W^{k+1}_j+B_jZ^{k+1})=\Lambda^k_j-\Lambda^{k+1}_j$. When $A_j$ is of full row rank,  then
$$\Lambda^{k+1}_j\in (A_jA^T_j)^{-1}\bullet \partial\|W^{k+1}_j\|_1, ~~\forall k\geq 1.$$
Obviously, for a fixed $j$, $\partial\|W^{k+1}_j\|_1$ is a compact set (\cite{rockafellar1970convex}), from which it follows that the sequence $\{\Lambda^k_j\}^\infty_{k=1}$ $(j=1,\cdots, J)$ is bounded (fixed $j$).
From the iteration procedure of Algorithm 1, we can see that

\begin{align*}
  &~~L_{\beta_k}(W^{k+1}_1, W^{k+1}_2, W^{k+1}_3,\cdots, W^{k+1}_J,Z^{k+1}, \Lambda^k_1,\cdots, \Lambda^k_J)\\
		&\leq L_{\beta_k}(W^k_1, W^{k+1}_2, W^{k+1}_3,\cdots, W^{k+1}_J,Z^{k+1}, \Lambda^k_1,\cdots, \Lambda^k_J)\\
		&\leq L_{\beta_k}(W^k_1, W^k_2, W^{k+1}_3,\cdots, W^{k+1}_J,Z^{k+1}, \Lambda^k_1,\cdots, \Lambda^k_J)\\
		& \vdots\\
		&\leq L_{\beta_k}(W^k_1,\cdots, W^k_J,Z^{k+1}, \Lambda^k_1,\cdots, \Lambda^k_J),\\
\end{align*}
and
\begin{align*}
&~~L_{\beta_k}(W^k_1,\cdots, W^k_J,Z^{k+1}, \Lambda^k_1,\cdots, \Lambda^k_J)\\
		&\leq L_{\beta_k}(W^k_1,\cdots, W^k_J,Z^k, \Lambda^k_1,\cdots, \Lambda^k_J)\\
		& = L_{\beta_{k-1}}(W^k_1,\cdots, W^k_J,Z^k, \Lambda^{k-1}_1,\cdots, \Lambda^{k-1}_J)\\
        &+\sum^J_{j=1}\langle \Lambda^{k-1}_j-\Lambda^k_j, A_j W^k_j+ B_j Z^k\rangle \\
        &+\frac{\beta_k-\beta_{k-1}}{2}\sum^J_{j=1}\|A_jW^k_j +B_j Z^k\|^2_F\\
        &= L_{\beta_{k-1}}(W^k_1,\cdots, W^k_J,Z^k, \Lambda^{k-1}_1,\cdots, \Lambda^{k-1}_J)\\
        &+ \frac{\beta_{k-1}+\beta_k}{2\beta^2_{k-1}}\sum^J_{j=1}\|\Lambda^k_j-\Lambda^{k-1}_j\|^2_F,
\end{align*}
where the last equality is achieved with the relation $\beta_{k-1}(A_jW^k_j +B_j Z^k) = \Lambda^{k-1}_j-\Lambda^k_j$.
Notice that $\{\Lambda^k_j\}^\infty_{k=1}$ is bounded and
$$\sum^\infty_{k=1} \frac{\beta_{k-1}+\beta_k}{2\beta^2_{k-1}}=\sum^\infty_{k=1} \frac{\rho^{k-1}\beta_0+\rho^k\beta_0}{2\rho ^{2(k-1)}\beta^2_0}=\frac {\rho(1+\rho)}{2(\rho-1)\beta_0}.$$
Thus, $L_{\beta_k}(W^{k+1}_1, \cdots, W^{k+1}_J,Z^{k+1}, \Lambda^k_1,\cdots, \Lambda^k_J)$ is upper bounded.
Moreover,
\begin{align*}
		\sum^J_{j=1} \|W^k_j\|_1
		&= L_{\beta_{k-1}}(W^k_1, \cdots, W^k_J,Z^k, \Lambda^{k-1}_1,\cdots, \Lambda^{k-1}_J)\\
&+\frac{1}{2\beta_{k-1}}\sum^J_{j=1}(\|\Lambda^{k-1}_j\|^2_F-\|\Lambda^k_j\|^2_F),
\end{align*}
is upper bounded,  which could be achieved by noticing the expression for $L_{\beta_{k-1}}$$(W^k_1$, $\cdots$, $W^k_J, Z^k$, $\Lambda^{k-1}_1$, $\cdots$, $\Lambda^{k-1}_J)$ and $\Lambda^{k-1}_j-\Lambda^k_j=\beta_{k-1} (A_j W^k_j+B_j Z^k)$.  Thus the sequence $\{W^k_j\}^\infty_{k=1}$ is bounded. Also notice that
$$A_j W^k_j+B_j Z^k=\frac{\Lambda^{k-1}_j-\Lambda^k_j}{\beta_{k-1}}\to 0 ~~\mbox {as}~~ k\to \infty.$$
		
Hence for any accumulation point $(W^*_1$, $\cdots$, $W^*_J$, $Z^*$, $\Lambda^*_1$, $\cdots$, $\Lambda^*_J$, $-\sum^J_{j=1}(B_jZ^*)^T\Lambda^*_j)$ of $\{(W^k_1$, $\cdots$, $W^k_J$, $Z^k$, $\Lambda^k_1$, $\cdots$, $\Lambda^k_J$, $-\sum^J_{j=1}(B_jZ^k)^T\Lambda^k_j)\}^\infty_{k=1}$, without any loss of generality, we can assume that  $(W^*_1$, $\cdots$, $W^*_J$, $Z^*$, $\Lambda^*_1$, $\cdots$, $\Lambda^*_J$, $-\sum^J_{j=1}(B_jZ^*)^T\Lambda^*_j)$ is the limit of  $\{(W^{k_i}_1$, $\cdots$, $W^{k_i}_J$, $Z^{k_i}$, $\Lambda^{k_i}_1$, $\cdots$, $\Lambda^{k_i}_J$,$-\sum^J_{j=1}(B_jZ^{k_i})^T\Lambda^{k_i}_j) \}^\infty_{{i}=1}$,  where $k_i$ is the subsequence of $k$, $i=1,\cdots,\infty$. Letting $i\to\infty$  and applying Lemma \ref{lemma3}, we have $A^T_{j} \Lambda^*_{1,j}\in \partial \|W^*_j\|_1,~~ j=1,\cdots,J$. Thus, $(W^*_1$, $\cdots$, $W^*_J$, $Z^*$, $\Lambda^*_1$, $\cdots$, $\Lambda^*_J$, $-\sum^J_{j=1}(B_jZ^*)^T\Lambda^*_j)$  satisfies the KKT conditions (8).
\end{proof}

\begin{proof}(Proof of Theorem 2)	
Since  $\lim_{k\to\infty}\frac{\Lambda^{k+1}_j-\Lambda^k_j}{\beta_k}=0$ and notice that $\frac{\Lambda^k_j-\Lambda^{k+1}_j}{\beta_k}=A_j W^{k+1}_j+B_jZ^{k+1}$, we have
$$\lim_{k\to\infty} A_j W^k_j+B_jZ^k=0.$$
For any accumulation point  $(W^*_1$, $\cdots$, $W^*_J$, $Z^*$, $\Lambda^*_1$, $\cdots$, $\Lambda^*_J$, $-\sum^J_{j=1}(B_jZ^*)^T\Lambda^*_j)$ of $\{(W^k_1$, $\cdots$, $W^k_J$, $Z^k$, $\Lambda^k_1$, $\cdots$, $-\sum^J_{j=1}(B_jZ^k)^T\Lambda^k_j)\}^\infty_{k=1}$,  there exists subsequence  $\{(W^{k_i}_1$, $\cdots$, $W^{k_i}_J$, $Z^{k_i}$, $\Lambda^{k_i}_1$,$\cdots$, $\Lambda^{k_i}_J$, $-\sum^J_{j=1}(B_jZ^{k_i})^T\Lambda^{k_i}_j)\}^\infty_{i=1}$ such that
$$\lim_{i\to\infty} W^{k_i}_j=W^*_j,~~~\lim_{i\to\infty} Z^{k_i}= Z^*, ~~~\lim_{i\to\infty} \Lambda^{k_i}_j=\Lambda^*_j.$$
Hence, for each fixed $j(j=1,\cdots, J)$,
\begin{align*}
&A_jW^*_j+B_j Z^*=\lim_{i\to\infty} A_jW^{k_i}_j+B_jZ^{k_i}=0,\\
&(Z^*)^TZ^*=\lim_{i\to\infty} (Z^{k_i})^TZ^{k_i}=I,
\end{align*}
which means $(W^*_1, \cdots, W^*_J,Z^*)$ is a feasible point of problem (7). Since $W^{k+1}_j$ solves problem (17), we have
\begin{align*}
&\frac{\beta_k}{\delta}\Big(W^k_j-\delta A^T_j (A_j W^k_j+B_j Z^{k+1}-\Lambda^k_j/\beta_k)-W^{k+1}_j\Big)\\
&\in \partial \|W^{k+1}_j\|_1,~~\forall k\geq 0
\end{align*}
or equivalently
$$A^T_j\Lambda^k_j+\Big(A^T_jA_j-\frac1\delta I\Big)\beta_{k-1}(W^k_j-W^{k-1}_j)\in \partial \|W^k_j\|_1,\forall k\geq 1,$$
by noticing
$$B_jZ^k=\frac{1}{\beta_{k-1}}(\Lambda^{k-1}_j-\Lambda^k_j)-A_jW^k_j.$$
Similarly, by passing to subsequence $\{k_i\}$, letting $i\to\infty$, and applying Lemma \ref{lemma3}, we can see that $A^T_j \Lambda^*_j\in \partial \|W^*_j\|_1$.
Therefore,  $(W^*_1$, $\cdots$, $W^*_J$, $Z^*$, $\Lambda^*_1$, $\cdots$, $\Lambda^*_J$, $-\sum^J_{j=1}(B_jZ^*)^T\Lambda^*_j)$ satisfies the KKT conditions (8). This completes the proof.
\end{proof}%
\section*{Acknowledgement}
The authors would like to thank Dr. Xiaowei Zhang from the Bioinformatics Institute, A*STAR,
Singapore, for discussing the theoretical analysis of the proposed algorithm.
	\ifCLASSOPTIONcaptionsoff
	\newpage
	\fi


	
	
	\bibliographystyle{IEEEtran}
	\bibliography{ADMMGCCA}
\end{document}